\pdfoutput=1
\documentclass[authoryear,3p]{elsarticle}

\usepackage{amsmath}
\usepackage{latexsym}
\usepackage{amssymb}
\usepackage{eufrak}
\usepackage{dsfont}
\usepackage{verbatim}
\usepackage{examplep}
\usepackage{graphicx}
\usepackage{color}
\usepackage{soul}
\usepackage{textcomp}
\usepackage{multirow}
\usepackage{xfrac}
\usepackage{array}
\usepackage{color}
\usepackage{soul}

\newcommand{\blue}[1]{\textcolor{blue}{#1}}
\newcommand{\white}[1]{\textcolor{white}{#1}}

\newcolumntype{L}[1]{>{\raggedright\let\newline\\\arraybackslash\hspace{0pt}}m{#1}}
\newcolumntype{C}[1]{>{\centering\let\newline\\\arraybackslash\hspace{0pt}}m{#1}}
\newcolumntype{R}[1]{>{\raggedleft\let\newline\\\arraybackslash\hspace{0pt}}m{#1}}

\biboptions{authoryear}

\usepackage[figuresright]{rotating}

\journal{Procedia Engineering}

\begin{document}

\title{Concurrent Pump Scheduling and Storage Level Optimization
    \\
    Using Meta-Models and Evolutionary Algorithms%
    \footnote{This article was presented in the 12th International Conference on Computing and Control for the Water Industry (CCWI'2013) and published on 04/23/2014 in Procedia Engineering. For citation, please use:
    \protect\\
    \protect\\
        \blue{Behandish, Morad and Wu, Zheng Yi, 2014. ``Concurrent Pump Scheduling and Storage Level Optimization Using Meta-Models and Evolutionary Algorithms.'' Procedia Engineering, 70, pp.103--112.
        }
    }
}

\author{Morad Behandish$^1$ and Zheng Yi Wu$^2$
    \\ {\small $^1$Department of Computer Science and Engineering, University of Connecticut, USA}
    \\ {\small $^2$Applied Research Group, Bentley Systems, Incorporated, USA}
}

\renewcommand{\today}{Technical Report No. BENT-TR-14-04, April 23, 2014}
\date{\today}

\begin{abstract}
    In spite of the growing computational power offered by the commodity hardware, fast pump scheduling of complex water distribution systems is still a challenge. In this paper, the Artificial Neural Network (ANN) meta-modeling technique has been employed with a Genetic Algorithm (GA) for simultaneously optimizing the pump operation and the tank levels at the ends of the cycle. The generalized GA+ANN algorithm has been tested on a real system in the UK. Comparing to the existing operation, the daily cost is reduced by about $10$$-$$15\%$, while the number of pump switches are kept below $4$ switches-per-day. In addition, tank levels are optimized ensure a periodic behavior, which results in a predictable and stable performance over repeated cycles.\\
\end{abstract}

\maketitle


\section{Introduction} \label{sec_intro}

Drinking water and waste water utilities account for about $3-4\%$ of the total energy use in the United States, and are responsible for more than $45$ million tons of greenhouse gas emission annually, as reported by the US Environmental Protection Agency (EPA). According to the same report, these systems account for $30-40\%$ of total energy consumption of municipal governments, and the energy-related operating costs are expected to increase as much as $20\%$ in the next fifteen years due to population growth and tightening drinking water regulations (\citet{EPA2012}).
These facts pronounce the increasing need for the water industry to improve water management strategies.

The problem of operation optimization is most frequently addressed by pump scheduling, i.e., predicting a set of either implicit control rules or explicit time-based specifications on when to turn pumps on and off, such that the supply or disposal service requirements are met with minimal energy cost. Pump scheduling has been extensively researched over the past few decades, using a variety of optimization techniques, the most popular of which has been Genetic Algorithms (GAs), including the studies by \citet{Mackle1995,Beckwith1995,Engelbrecht1996,Nitivattananon1996,DeSchaetzen1998,Wu2001,Kelner2003} and \citet{Van2004}. The earliest studies were mostly conducted based on a single objective function (i.e., operation energy or cost). Multi-objective evolutionary optimization algorithms have also been used by \citet{Savic1997,Sotelo2002,Baran2005,Lopez2005}, and \citet{Wang2009}. In either case, pump scheduling with the direct application of hydraulic solvers is computationally intensive when applied to models of large utilities. To overcome this difficulty, parallel computing technology has been utilized by \citet{Von2004,Wu2009a} and \citet{Wu2012,Wu2012a} to speed up the optimization.
The speed-up factors scale well by increasing the number of physical computing cores; however, expensive multiprocessor systems are required, especially for real-time applications demanding continuous updates of the pump control policies.

In addition to the pump schedule, there are other hydraulic parameters that are indirectly related to the energy consumption. One such set of parameters are the operation ranges of storage tank levels, which can introduce an independent set of additional decision variables alongside pump control settings.
In real systems, a tank is not necessarily designed to result in optimal energy consumption when its entire capacity is utilized, nor is necessarily placed at optimal locations. As a result, some tanks might have more impacts on energy saving than the others if they are filled when electricity is inexpensive and drained during the peak-demand periods.

Different techniques have been used for hydraulic system modeling, ranging from empirical models to simplified hydraulic models and state-of-the-art hydraulic simulation packages, as reviewed by \citet{Rao2007}.
An alternative effective solution to reduce the computation time is offered by machine learning. Artificial Neural Networks (ANNs) have been extensively utilized for predicting the hydraulic state of water distribution and disposal systems in the recent years by \citet{Broad2005} and \citet{Tabach2007}, and for the specific problem of real-time pump operation control by \citet{Jamieson2007,Rao2007}, and \citet{Rao2007a}. These techniques were applied to real systems in \citet{Salomons2007,Martinez2007,Shamir2008} and \citet{Behandish2012}. Furthermore, the ANN training process lends itself well to high-throughput parallel processing on the Graphics Processing Unit (GPU), as demonstrated by \citet{Wu2011a} and \citet{Behandish2012}.
The multi-ANN meta-modeling was later improved and generalized by \citet{Behandish2013} to encompass a wider spectrum of systems with diverse state variable combinations to be predicted with significantly improved accuracy and robustness.

\begin{figure}
    \centering
    \includegraphics[width=\textwidth]{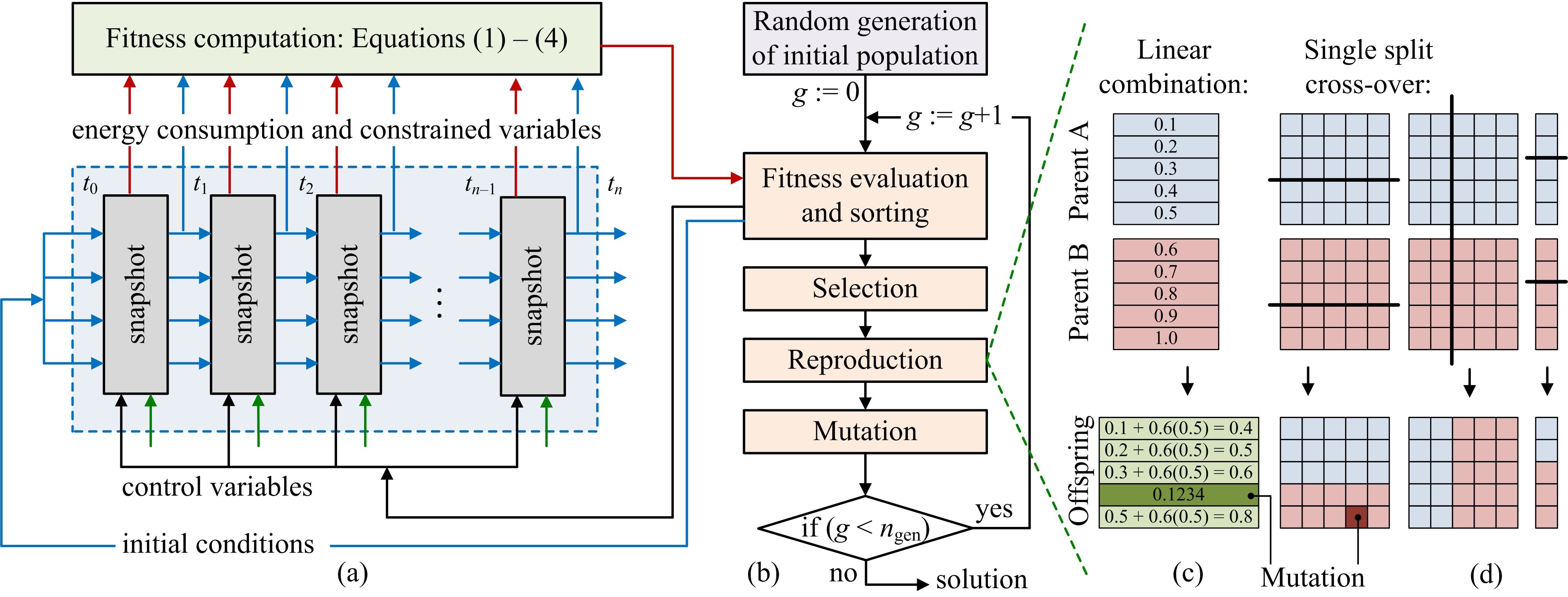}
    \caption{Evolutionary optimization of water distribution systems: (a) the multi-ANN meta-model used to predict the hydraulic response; (b) the Genetic Algorithm (GA) used to optimize the pump operation schedule; (c) and (d) reproduction operators of the GA.} \label{figure1}
\end{figure}

In this article, the generalized multi-ANN meta-modeling technique developed by the authors is utilized in combination with a modified Genetic Algorithm (GA).
The modified GA is employed to search for the decision space made of both discrete variables (e.g., pump/valve statuses) and continuous variables (e.g., tank levels). The results on a case study system are provided and compared with those of the earlier studies in terms of energy saving, number of required pump switches, and periodic tank water level variations.

\section{Generalized Meta-Modeling} \label{sec_meta}

\begin{figure}
    \centering
    \includegraphics[width=\textwidth]{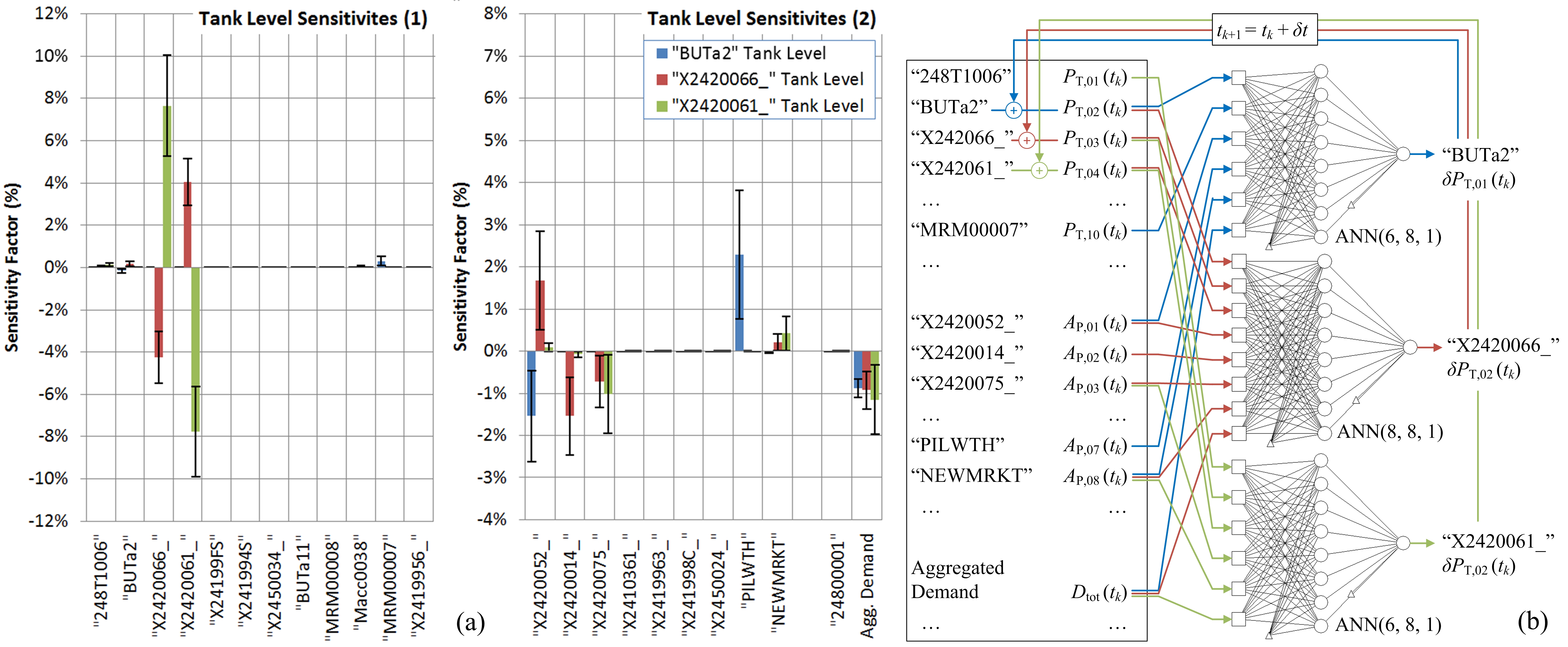}
    \caption{Multi-ANN meta-model construction: (a) sensitivity analysis results; (b) sample sub-ANN structures to predict the hydraulic response.} \label{figure2}
\end{figure}

Fig. \ref{figure1} (a) is a schematic illustration of the Extended Period Simulation (EPS) in a succession of time, with each snapshot being replicated by a set of independent neural networks (\citet{Behandish2013}). The initial conditions such as the storage tank levels are specified at time $t_0$, while control variables including pump and valve statuses (on or off), pump speed settings (when applicable), etc., are explicitly controlled for each time interval. In addition, time patterns are pre-specified for nodal demands, electricity tariffs, reservoir heads, etc. Each snapshot of the simulation relates the state variables at time $t_i$ to those at $t_{i+1} = t_i + \delta t$, where typically $\delta t = 1$ hour. Before setting up such a network, a sensitivity analysis was carried out to find out which state variables to select as the inputs of each sub-ANN to predict a particular output (\citet{Behandish2013}). Fig. \ref{figure2} (a) shows the results of the sensitivity analysis for a Demand Monitoring Zone (DMZ) system in the UK. The corresponding sub-ANN constructions are depicted in Fig. \ref{figure2} (b), where the inputs with the highest impacts on each output are maintained. Therefore, 12 sub-ANN structures are formed based on these input/output couplings, together with an additional sub-ANN that is used to predict the energy consumption rates from the pump statuses. The sub-ANNs are trained one at a time or in parallel on the GPU with NVIDIA’s Compute Unified Device Architecture (\citet{CUDA2012}), using large datasets obtained with the hydraulic model.
The reader is referred to \citet{Behandish2013} for further details on the generalized meta-modeling.

\section{Optimization Model} \label{sec_prob}

The trained and verified multi-ANN meta-model was integrated with an evolutionary optimization algorithm to evaluate the fitness of a trial solution for the system operation control, and to search for a near-optimal solution.
Two distinct sets of decision variables are considered in general:

\begin{enumerate}
    \item {\it Control variables}: The control variables include the statuses and settings of a control element (i.e., pump or valve) at every control interval. The statuses are defined as the fraction of the time interval $\delta t$ during which the pump or valve has been kept open (e.g., 0 for closed, 1 for open) and the settings can represent pump speeds, valve flow/pressure settings, etc., as detailed by \citet{Behandish2013}.

    \item {\it Initial Conditions}: The initial storage tank levels, which are typically constrained to be recovered at the end of the operation cycle, are also included in the decision variable set. The tank levels at $t = t_0$ are decided by the optimizer, and those at $t > t_0$ are predicted successively by the meta-model.
\end{enumerate}

Using the convention of \citet{Behandish2013}, all state variables are normalized with the following linear equation, to make them dimensionless and with similar orders of magnitude:

\begin{equation}
    \bar{s}(t) = \frac{s_i(t)~-s_{i,{\rm min}}} {s_{i,{\rm max}} - s_{i,{\rm min}}}, \quad i=1,2,\cdots,n.
\end{equation}
The first set of decision variables can be represented in two matrices, one for statuses and the other for settings. The former is a $(n_{\rm P} + n_{\rm V}) \times m$ matrix $[A_{i, k}]$ made of $n_{\rm P}$ pump statuses $0 \leq A_{{\rm P}, i} (t_k) \leq 1 ~(1 \leq i \leq n_{\rm P})$, and $n_{\rm V}$ valve statuses $0 \leq A_{{\rm V}, i} \leq 1 (t_k) ~(1 \leq i \leq n_{\rm V})$ over $m$ control intervals starting at $t_k = t_0 + k \delta t ~(0 \leq k \leq m - 1)$. Similarly, the latter is a $(n'_{\rm P} + n'_{\rm V}) \times m$ matrix $[\bar{S}_{i, k}]$ that represents $n'_{\rm P}$ normalized pump settings $0 \leq \bar{S}_{{\rm P}, i} (t_k) \leq 1 ~(1 \leq i \leq n'_{\rm P})$, and $n'_{\rm V}$ normalized valve settings $0 \leq \bar{S}_{{\rm V}, i}(t_k) \leq 1 ~(1 \leq i \leq n'_{\rm V})$ defined over the same $m$ control intervals. The second set of decision variables, on the other hand, can be represented by a vector $[\bar{P}_{j,0}]$ made of $n_{\rm T}$ normalized tank levels $\bar{P}_{{\rm T}, j} (t_0)$ at the beginning of the cycle.

As schematically depicted in Fig. \ref{figure1} (b), the Genetic Algorithm (GA) generates the trial solution set ${\bf S}_{\rm GA}$ of the aforementioned $n = (n_{\rm P} + n_{\rm V} + n'_{\rm P} + n'_{\rm V}) \times m + n_{\rm T}$ variables.
The trial solution together with other parameters $S_{\rm par}$, that may include the demand patterns, the reservoir head patterns, the pre-specified pump and valve settings, etc., all in normalized forms, are passed to the multi-ANN meta-model. For each GA trial, the extended period simulation is replicated by the successive calls to the trained ANNs. The ANN outputs may include the pump energy rates, the tank levels, and possibly other dependent state variables such as the junction pressures and the pipe flow rates. Part of the output is passed to the fitness computing routine, where the object function and penalty function are evaluated.

The objective function is defined as the total pumping energy cost, normalized in the following form:

\begin{equation} \label{eq_1}
    \bar{F}({\bf S}_{\rm GA}; {\bf S}_{\rm par}) = \frac{1}{m} \sum_{k = 0}^{m -1} \bar{C}(t_k) \bar{E}_{\rm tot}(t_k),
\end{equation}
where $\bar{C}(t) = C(t) / C_{\rm max}$ is the normalized electricity tariff, $\bar{E}_{\rm tot}(t) = E_{\rm tot}(t) / E_{\rm max}$ is the normalized energy consumption rate aggregated over all pumps, $\Delta t = t_{\rm opt} - t_0$ is the operation time, $\delta t = \Delta t / m$ is the control interval, and $t_k = t_0 + k \delta t$ is the discrete time step $0( \leq k \leq m - 1)$.

The water distribution service requirements are quite diverse among different systems. In this article, three different classes of generalized constraints are defined and implemented as follows:

\begin{enumerate}
    \item {\it Time-Based Constraints}: The time-based constraints specify all of the requirements on the selected hydraulic state responses over the control horizon, ranging from the tank levels and the junction pressures to the pipe flow rates, etc. These constraints are typically expressed as $s_{i,{\rm min}}(t) \leq s_i(t) \leq s_{i,{\rm max}}(t)$, where $s_i(t)$ is any state variable that is dependent on the decision variables, and $s_{i,{\rm min}}(t)$ and $s_{i,{\rm max}}(t)$ are the prescribed lower-bound and upper-bound. The most common time-based constraints are:
        \begin{enumerate}
            \item Lower-bounds on tank levels $P_{{\rm T}, j}(t) \geq P_{\rm T, min}$, e.g., $P_{\rm T, min} = 30\%$ of capacity, maintained for emergency.
            \item Upper-bounds on tank levels $P_{{\rm T}, j}(t) \leq P_{\rm T, max}$, e.g., $P_{\rm T, max} = 95\%$ of capacity, to avoid water overtopping.
            \item Requirements on storage at early morning hours $P_{{\rm T}, j} (t) \geq P_{\rm T, AM} (t)$, e.g., $P_{\rm T, AM}(t) = 80\%$ of capacity at early morning $t = $ 6:00 to 8:00 AM, and $P_{\rm T, AM}(t) = -\infty$ at all other times (i.e., no lower-bound).
            \item Lower-bounds on node pressures $P_{{\rm J}, j}(t) \geq P_{\rm J, min}$, e.g., $P_{\rm J, min} = $ minimum pressure required at consumer end.
            \item Upper-bounds on node pressures $P_{{\rm J}, j}(t) \leq P_{\rm J, max}$, e.g., $P_{\rm J, max} = $ maximum pressure required to avoid leakage, or the maximum pressure that the junctions can endure, whichever is smaller.
            \item Lower-bounds on pipe flow rates $Q_{{\rm I}, j}(t) \geq Q_{\rm I, min}$, e.g., $Q_{\rm I, min} = $ minimum flows required to avoid stagnation and maintain water quality.
            \item Upper-bounds on pipe flow rates $Q_{{\rm I}, j}(t) \leq Q_{\rm I, max}$, e.g., $Q_{\rm I, max} = $ maximum flows that pipes can endure.
        \end{enumerate}
    \item {\it Periodicity Constraints}: Regardless of the time-variant constraints on the values, some state variables are constrained to return to their initial values at the end of the cycle as an operational requirement, so that the operation would repeat itself periodically under similar conditions at the subsequent cycles. For instance, a tank level at the end of the cycle can be constrained to be in a tolerance range of its initial water level, e.g., $| P_{{\rm T}, j} (t_{\rm opt}) - P_{{\rm T}, j} (t_{\rm opt}) | \leq \Delta P_{{\rm T},j}$.

    \item {\it Switch Constraints}: A pump scheduling scenario is of practical significance only if the number of pump switches per day (i.e., the number of times that each pump is turned on and off) is restricted, e.g., to $4$ or $6$ switches per day for each pump. This is because numerous pump switches are detrimental to the pump's life-cycle, resulting in prohibitively large maintenance and replacement costs.
\end{enumerate}
The violation of each inequality constraint written in the standard form of $g(\cdot) \leq 0$, is quantified with $\langle g(\cdot) \rangle := \min\{0, g(\cdot)\}$. For a trial solution ${\bf S}_{\rm GA}$, the violation measured over the cycle $\Delta t = t_{\rm opt} - t_0$ is formulated as:
\begin{align}
    \bar{G}({\bf S}_{\rm GA}; {\bf S}_{\rm par}) &= \sum_{i=1}^{n} \frac{c_{1,i}}{m} \sum_{k = 0}^{m-1} \left[ \left\langle \bar{s}_i(t_k) - \bar{s}_{{\rm max},i}(t_k) \right\rangle + \left\langle \bar{s}_{{\rm min},i}(t_k) - \bar{s}_i (t_k) \right\rangle \right],  \label{eq_2} \\
    &+ \sum_{j=1}^{n} c_{2,j} ~\left\langle \Delta \bar{s}_{j}|_0^{t_{\rm opt}} - \Delta \bar{s}_{{\rm max},j} \right\rangle + \sum_{p = 1}^{n_{\rm P}} c_{3,p} ~\left\langle \eta_{p}|_{0}^{t_{\rm opt}} - \eta_{{\rm max},p} \right\rangle. \label{eq_2}
\end{align}
where $|\Delta \bar{s}_{j}|_0^{t_{\rm opt}} := |\bar{s}_{j}(t_{\rm opt}) - \bar{s}_{j}(t_0)|$, and $\eta_{p}|_{0}^{t_{\rm opt}}$ is the actual number of pump switches per cycle, e.g., $24$ hours. The first term on the right is the sum of violations of the time-base constraints, integrated for each state variable over the time interval $\Delta t = t_{\rm opt} - t_0$. The second term measures the violation of the periodicity requirements with the normalized tolerances $\Delta \bar{s}_{{\rm max},j}$, and the last term quantifies the violation of pump switch constraints with the maximum allowable number of switches $\eta_{{\rm max},p}$. The state variables $\bar{s}_i(t)$ in the first two terms can be the normalized tank levels $\bar{s}_i(t) := \bar{P}_{{\rm T}, i}(t)~(1 \leq i \leq n_{\rm T})$, the normalized junction pressures $\bar{s}_i(t) := \bar{P}_{{\rm J}, i}(t)~(1 \leq i \leq n_{\rm J})$, the normalized pipe flow rates $\bar{s}_i(t) := \bar{Q}_{{\rm I}, i}(t)~(1 \leq i \leq n_{\rm I})$, etc. The weight factors $c_{1,i}$, $c_{2,j}$, and $c_{3,p}$ are set to 1 by default for constrained variables, and to 0 for the unconstrained variables.

The objective function and the violation function formulated in Eqs. (\ref{eq_1}) and (\ref{eq_2}) are combined using additive penalty method into the following penalized objective function:
\begin{equation} \label{eq_4}
    \bar{F}^{\ast}({\bf S}_{\rm GA}; {\bf S}_{\rm par}) = \bar{F}({\bf S}_{\rm GA}; {\bf S}_{\rm par}) + \mathcal{P} \times \bar{G}({\bf S}_{\rm GA}; {\bf S}_{\rm par}),
\end{equation}
where $\mathcal{P}$ is the penalty factor, typically selected in the order of $\mathcal{P} \sim 10^2-10^3$ depending on how strictly the constraints are being enforced. The penalized objective function $F^{\ast}({\bf S}_{\rm GA}; {\bf S}_{\rm par})$ is used as a measure of fitness of the trial decision set ${\bf S}_{\rm GA}$ generated by the GA. The lower the value of this function is, the fitter the trial scenario is, hence more likely it is to survive or pass its properties to the next generations of the evolutionary optimization.

\section{Optimization Algorithm} \label{sec_GA}

The optimization problem is solved by combining the classic binary operators with those of a modified Genetic Algorithm (GA) (\citet{Mundo2007}). The hybrid search algorithm is schematically illustrated in Fig. \ref{figure1} (b).
The algorithm iterates over $n_{\rm gen}$ generations, and at each generation $g$ where $0 \leq g \leq n_{\rm gen} - 1$, a population of $n_{\rm pop}$ individuals (i.e., chromosomes) are maintained. Each individual is represented by a set of normalized decision variables ${\bf S}_{{\rm GA}, g}$ composed of the control variables (i.e, the matrices of pump/valve statuses and normalized settings) and initial conditions (i.e., the vectors of normalized initial storage levels) as explained in Section \ref{sec_prob}. The normalized decision variables ${\bf S}_{{\rm GA}, 0}$ of the initial population are randomly assigned with values in $[0, 1]$, corresponding to random values in the physical domain in each state variable's min/max range. The subsequent generations ${\bf S}_{{\rm GA}, g} (1 \leq g \leq n_{\rm gen} - 1)$ are descended from their parents in the previous generation ${\bf S}_{{\rm GA}, g-1}$, through a series of evolutionary operations. The modified GA utilizes a combination of evolutionary operators that are designed for binary-coded chromosomes (e.g., for pump/valve statuses that are constrained to a small number of switches) as well as real-valued variables (e.g., tank levels or pump speed settings, if applicable) without a need to use binary-coding for the latter.

At the beginning of each iteration, the fitness of the individuals are computed using the multi-ANN meta-model and sorted in descending order of the penalized objective function $F^{\ast}({\bf S}_{\rm GA}; {\bf S}_{\rm par})$. The algorithm follows an elitist strategy, hence a constant fraction $f_{\rm elit} \sim 1 - 5 \%$ of the population consisting of the fittest individuals are directly sent to the next generation. Another fraction $f_{\rm rand} \sim 5 - 15\%$ of the population is randomized at every generation, to enhance the exploration and avoid local minima. The rest of the population consisting of $1 - (f_{\rm elit} + f_{\rm rand}) \sim 80 - 95 \%$ are generated from reproduction (i.e., breeding) operations. The selection of the parents for reproduction can be based on different probability distributions. In this algorithm, the method of normalized geometric ranking selection is used (\citet{Mundo2007}), in which the probability of an individual to be selected is calculated from its rank in the sorted array based on fitness:
\begin{equation} \label{eq_5}
    P_{\rm rep}(r) = \frac{P_0 (1-P_0)^r}{1-(1-P_0)^{n_{\rm pop}}}, \quad r = 0, 1, \cdots, n_{\rm pop} - 1,
\end{equation}
where $0 < P_0 < 1$ is a constant and is proportional to the probability of selecting the fittest individual, $r$ is the rank of the individual ($r = 0$ for the fittest individual, and $r = n_{\rm pop} - 1$ for the least fit individual of that generation).

Once the parents are selected, different reproduction operators are employed to produce one offspring at a time:

\begin{enumerate}
    \item {\it Linear Combination}: With a probability of $0 \leq P_{\rm com} \leq 1$, the offspring genes are generated using a linear combination of the parents' genes:

        \begin{equation}
            \bar{s}_{{\rm off}, i} = \bar{s}_{{\rm A}, i} + R_{\rm com} (\bar{s}_{{\rm B}, i} - \bar{s}_{{\rm A}, i}), \quad R_{\rm com} \in \big[-\epsilon, 1+\epsilon \big),
        \end{equation}
        where the subscripts ``${\rm off}$'', $A$, and $B$ refer to the offspring and the two parents, respectively (\citet{Mundo2007}), as exemplified in Fig. \ref{figure1} (c) with $R_{\rm com} = 0.6$. The combination factor $R_{\rm com}$ is a random number in $[-\epsilon, 1+\epsilon)$ where $\epsilon \sim 0-0.2$ is the overshoot. It is evident that this reproduction scheme applies to the real-valued decision variables only. Therefore, for the pump/valve statuses restricted to binary values, the linear combination is replaced with a simple random selection, i.e., for every gene $\bar{s}_{{\rm off}, i} := \bar{s}_{{\rm A}, i}$, or $\bar{s}_{{\rm off}, i} := \bar{s}_{{\rm B}, i}$, each with $50\%$ probability.
    \item {\it Single Split Cross-Over}: With a probability of $0 \leq P_{\rm crs} \leq 1$, the offspring genes are generated from a single point cross-over operation on the parents. For the pump/valve statuses and settings arranged into matrices, the splitting can occur along the rows or columns, each with a $50\%$ probability. For the initial conditions arranged into a vector, on the other hand, it is a simple splitting along the one-dimensional array, as illustrated in Fig. \ref{figure1} (d).
    \item {\it Direct Transfer}: With a probability $1 - (P_{\rm com} + P_{\rm crs})$, one of the two parents are directly transferred to the next generation, each with $50\%$ probability. This could be replaced with an elitist procedure, in which the fitter parent is more likely to be transferred.
\end{enumerate}
With the above operations, one offspring is created by a pair of parents. The selection of the parents and the reproduction have to be repeated $1 - (f_{\rm elit} + f_{\rm rand})$ times to generate enough number of offsprings for the next generation.

The produced offsprings are subjected to mutation with a small probability $P_{\rm mut} \sim 1-2\%$ to avoid local minima and promote global exploration of the search landscape. For the real-valued decision variables, a uniform random mutation operation is employed, which means a reassignment of one of the normalized decision variables into a random value in $[0, 1]$, corresponding to a random value in the physical min/max range, as illustrated in Fig. \ref{figure1} (c). For binary-coded decision variables, on the other hand, the mutated gene is changed from $0$ to $1$ and from $1$ to $0$, as shown in Fig. \ref{figure1} (d).

As illustrated in Fig. \ref{figure1} (b), once the next generation of $n_{\rm pop} \sim 100-300$ new individuals are produced, the procedure is repeated for enough number of generations $n_{\rm gen} = 5,000$, until a near-optimal set of control variables and initial tank levels is found. In addition to the selection, reproduction, and mutation, every several generations $n_{\rm res} \sim 100-500$, the entire population except the elite fraction are completely reset to random decision variables to simulate several optimization sessions following one another, always keeping the best results obtained so far.

\section{Results \& Discussion} \label{sec_res}

The generalized GA+ANN technique was applied to a Demand Monitoring Zone (DMZ) in the UK (\citet{Wu2009}). The hydraulic model for this system is composed of $3,537$ junctions, $3,273$ pipes, $5$ reservoirs, $12$ storage tanks, $19$ constant-speed pumps, and $420$ valves.
The optimization was carried out over $\Delta t = 24$ hours with time steps $\delta t = 1$ hour. Two different optimization scenarios are reported here, one with pump scheduling of $9$ active pumps (the same pumps utilized in the existing operation), which involves $(9 \times 24) = 216$ decision variables, and the other with simultaneous pump scheduling and storage optimization of all $12$ tanks, which gives rise to $216 + 12 = 238$ decision variables. For this particular case study, the three categories of constraints defined in Section \ref{sec_prob} are specified, including: (1) for $10$ out of $12$ tank levels, $P_{{\rm T},j} \geq 30\%$ of capacities for emergency, and $P_{{\rm T},j} \leq 95\%$ of capacities to avoid overtopping; (2) the periodicities of the same tank levels are enforced as $|P_{{\rm T},j} (24) - P_{{\rm T},j}(0)| \leq 10 {\rm cm}$, which is less than $2.5\%$ of the smallest tank capacity. (3) For every active pump, a maximum of $4$ switches per $24$ hours are allowed. No explicit constraints on junction pressures or pipe flow rates are specified. The GA parameters are set as: $\mathcal{P} = 1,000$, $n_{\rm gen} = 5,000$, $n_{\rm res} = 100$, $n_{\rm pop} = 300$, $f_{\rm elit} = 0.01$, $f_{\rm rand} = 0.10$, $P_0 = 0.05$, $P_{\rm com} = 0.40$, $P_{\rm crs} = 0.50$, and $P_{\rm mut} = 0.01$.

\begin{figure}
    \centering
    \includegraphics[width=\textwidth]{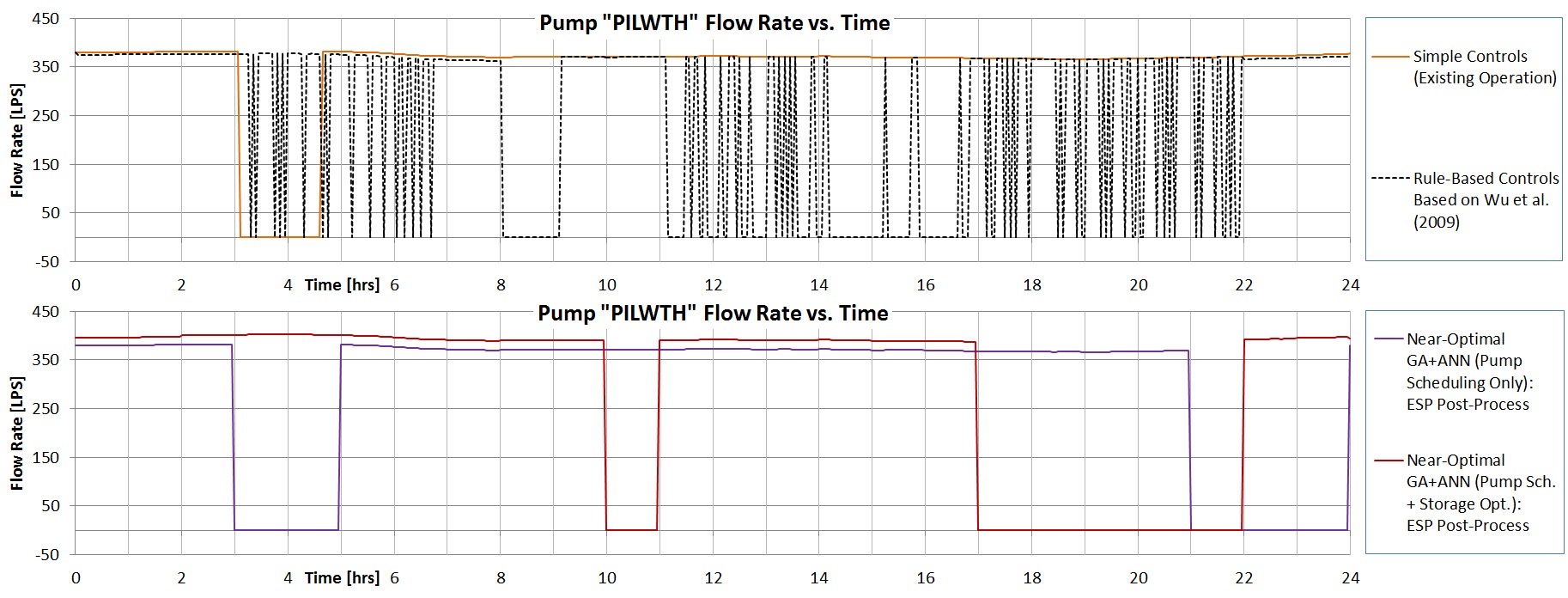}
    \caption{Pump flows over 24 hours for 4 operation scenarios. The results are obtained with EPANET using a simulation time step of 5 minutes.} \label{figure3}
\end{figure}

The GA+ANN solutions are compared with the existing operation and the pump scheduling study by \citet{Wu2009} on the same system. The existing operation uses simple controls that define when the pumps are turned on and off based on tank levels; for instance, the pump ``PILWTH'' is turned on if the tank ``BUTa2'' level falls below 5.30 meters, and turned off if the level starts to exceed 5.73 meters. The thresholds are based on experience and does not necessarily ensure cost-effective operation. The study by \citet{Wu2009}, on the other hand, used rule-based controls, but additional rules were specified with greater thresholds on the minimum storage at night when the electricity is inexpensive; e.g., at clock-times between 10:00 PM and 8:00 AM, the pump ``PILWTH'' is turned on if tank ``BUTa2'' level falls below 5.73 meters, and turned off otherwise. Both simple and rule-based controls secure the controlled tank levels between the limits, but they rely on the known relationships of pumps with tanks, and do not guarantee any bound on the number of pump switches that is required for this purpose. The cost can be significantly reduced by optimizing the rules in the latter method, but as shown in Table \ref{tab_1}, this comes at the expense of numerous pump switches enforced by the control rules. For instance, the utilization of pump ``PILWTH'' is decreased by about $60\%$ with the optimized rules, saving around \textlira 190 per day. However, as Fig. \ref{figure3} shows, the reduced pump utilization is made possible with numerous pump switches at high frequencies to satisfy the rules, which not desirable in practice.

\begin{table} \scriptsize
    \caption{Comparison of pump utilization, energy cost, number of pump switches (\#PS), and tank level periodicities for 4 operation scenarios. 
    }
    \begin{tabular}{| L{1.85cm} | R{0.85cm} | L{0.85cm} | R{0.5cm} | R{0.85cm} | L{0.85cm} | R{0.5cm} | R{0.85cm} | L{0.85cm} | R{0.5cm} | R{0.85cm} | L{0.85cm} | R{0.5cm} |}
    \hline
    & \multicolumn{3}{ c| }{Simple Controls} & \multicolumn{3}{ c| }{Rule-Based Controls} & \multicolumn{3}{ c| }{Near-Optimal GA+ANN} & \multicolumn{3}{ c| }{Near-Optimal GA+ANN} \\
    {\bf Case Studies} $\rhd$ & \multicolumn{3}{ c| }{(Existing Operation)} & \multicolumn{3}{ c| }{(Based on \citet{Wu2009})} & \multicolumn{3}{ c| }{(Pump Scheduling Only)} & \multicolumn{3}{ c| }{(Pump Sch. + Storage Opt.)} \\ \hline
    \hline
    {\bf Pump ID} & {\bf Util.} & {\bf Cost} & {\bf \#PS} & {\bf Util.} & {\bf Cost} & {\bf \#PS} & {\bf Util.} & {\bf Cost} & {\bf \#PS} & {\bf Util.} & {\bf Cost} & {\bf \#PS} \\ \hline
    X2420052\_ & 100$~$\% & \textlira 226.3 & 0 & 100$~$\% & \textlira 179.1   & 0  & 79.2\% & \textlira 178.1 & 4   & 83.3\% & \textlira 183.5 & 4 \\ \hline
    X2420014\_ & 100$~$\% & \textlira 350.4 & 0 & 41.7\%   & \textlira 127.2   & 28 & 41.7\% & \textlira 127.2 & 2   & 37.5\% & \textlira 122.6 & 4 \\ \hline
    X2420075\_ & 0.00\%   & \textlira \white{0}0.00  & 0 & 21.5\%   & \textlira 51.48   & 26 & 37.5\% & \textlira 192.3 & 4   & 41.7\% & \textlira 173.7 & 4 \\ \hline
    X2410361\_ & 36.0\%   & \textlira 24.39 & 2 & 41.7\%   & \textlira 22.54   & 3  & 58.3\% & \textlira 39.41 & 4   & 12.5\% & \textlira \white{0}6.36 & 4 \\ \hline
    X2419963\_ & 36.0\%   & \textlira 24.39 & 2 & 41.7\%   & \textlira 22.54   & 3  & 41.7\% & \textlira 27.34 & 2   & 75.0\% & \textlira 47.81 & 4 \\ \hline
    X241998C\_ & 13.6\%   & \textlira 04.14 & 2 & 26.8\%   & \textlira \white{0}4.23    & 48 & 25.0\% & \textlira \white{0}5.69 & 4   & 25.0\% & \textlira \white{0}5.67 & 2 \\ \hline
    X2450024\_ & 52.7\%   & \textlira 46.79 & 3 & 23.0\%   & \textlira 15.21   & 56 & 37.5\% & \textlira 34.80 & 4   & 37.5\% & \textlira 27.53 & 2 \\ \hline
    PILWTH     & 93.6\%   & \textlira 276.6 & 2 & 36.0\%   & \textlira 84.91   & 116& 79.2\% & \textlira 236.3 & 4   & 75.0\% & \textlira 211.7 & 4 \\ \hline
    NEWMRKT    & 0.00\%   & \textlira \white{0}0.00  & 0 & 41.7\%   & \textlira 177.6   & 3  & 12.5\% & \textlira 59.48 & 2   & 8.33\% & \textlira 32.89 & 2 \\ \hline
    {\bf Total} & N/A      & \textlira 953.0 & N/A & N/A    & \textlira 684.8   & N/A & N/A   & \textlira 900.7 & N/A & N/A& \textlira 821.8 & N/A \\ \hline
    \hline
    {\bf Cost Savings} & \multicolumn{3}{ r| }{ N/A } & \multicolumn{3}{ r| }{ \textlira 268.2 (28.15\%) } & \multicolumn{3}{ r| }{ \textlira 52.36 (5.49\%) } & \multicolumn{3}{ r| }{ \textlira 131.29 (13.78\%) } \\ \hline
    {\bf No. Switches} & \multicolumn{3}{ r| }{ $\eta_{\rm max} < 4$ per day (\checkmark) } & \multicolumn{3}{ r| }{ $\eta_{\rm max} \gg 4$ per day (\ding{53}) } & \multicolumn{3}{ r| }{ $\eta_{\rm max} =4$ per day (\checkmark) } & \multicolumn{3}{ r| }{ $\eta_{\rm max} = 4$ per day (\checkmark) } \\ \hline
    {\bf Periodicities} & \multicolumn{3}{ r| }{ $|\Delta_{\rm max}| \gg 10.0$ cm (\ding{53}) } & \multicolumn{3}{ r| }{ $|\Delta_{\rm max}| \gg 10.0$ cm (\ding{53}) } & \multicolumn{3}{ r| }{ $|\Delta_{\rm max}| = 15.3$ cm (\ding{53}) } & \multicolumn{3}{ r| }{ $|\Delta_{\rm max}| = 10.3$ cm (\checkmark) } \\ \hline
    \end{tabular} \label{tab_1}
\end{table}

Table \ref{tab_1} compares the pump utilization characteristics of the 4 different operations, obtained with EPANET using a time step of $5$ minutes for every one of the 4 scenarios. It is observed that the rule-based control strategy saves about $30\%$ of the daily cost, but the number of pump switches is prohibitively large (see Fig. \ref{figure3}). The near-optimal GA+ANN solution, on the other hand, can results in $10-15\%$ cost saving with a maximum of 4 pump switches per day. Furthermore, both of the existing simple and rule-based controls result in large differences between the initial and the final tank levels, making it difficult to anticipate the operating cost and performance of subsequent cycles.
The GA+ANN solution, on the other hand, guarantees repeatability of the operation by bringing the tank levels back to the initial conditions at the end of each cycle. It is also worthwhile noting that the existing simple and rule-based controls are on $7$ out of $12$ tanks, hence emptied or overtopped levels are observed for the other $5$ tanks; while GA+ANN studies constrain $10$ tanks to be periodic.

\begin{figure}
    \centering
    \includegraphics[width=\textwidth]{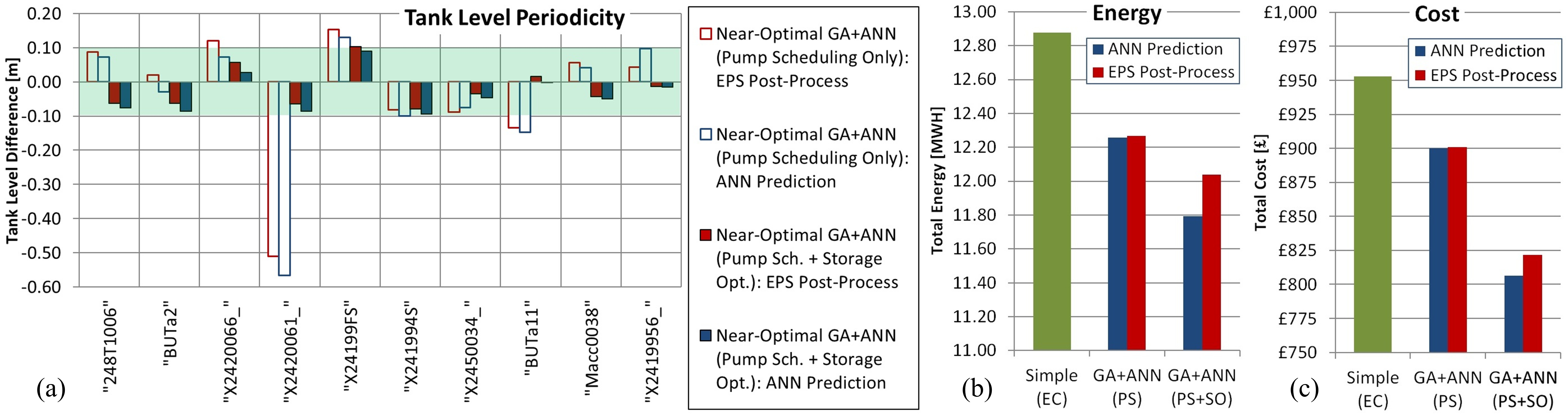}
    \caption{GA+ANN optimization results: (a) tank level difference between the two ends of cycle; (b) comparison of the energy consumptions; and (c) comparison of the energy costs. The post-processing results are generated with EPANET using a simulation time step of $1$ hour.} \label{figure4}
\end{figure}

\begin{figure}
    \centering
    \includegraphics[width=\textwidth]{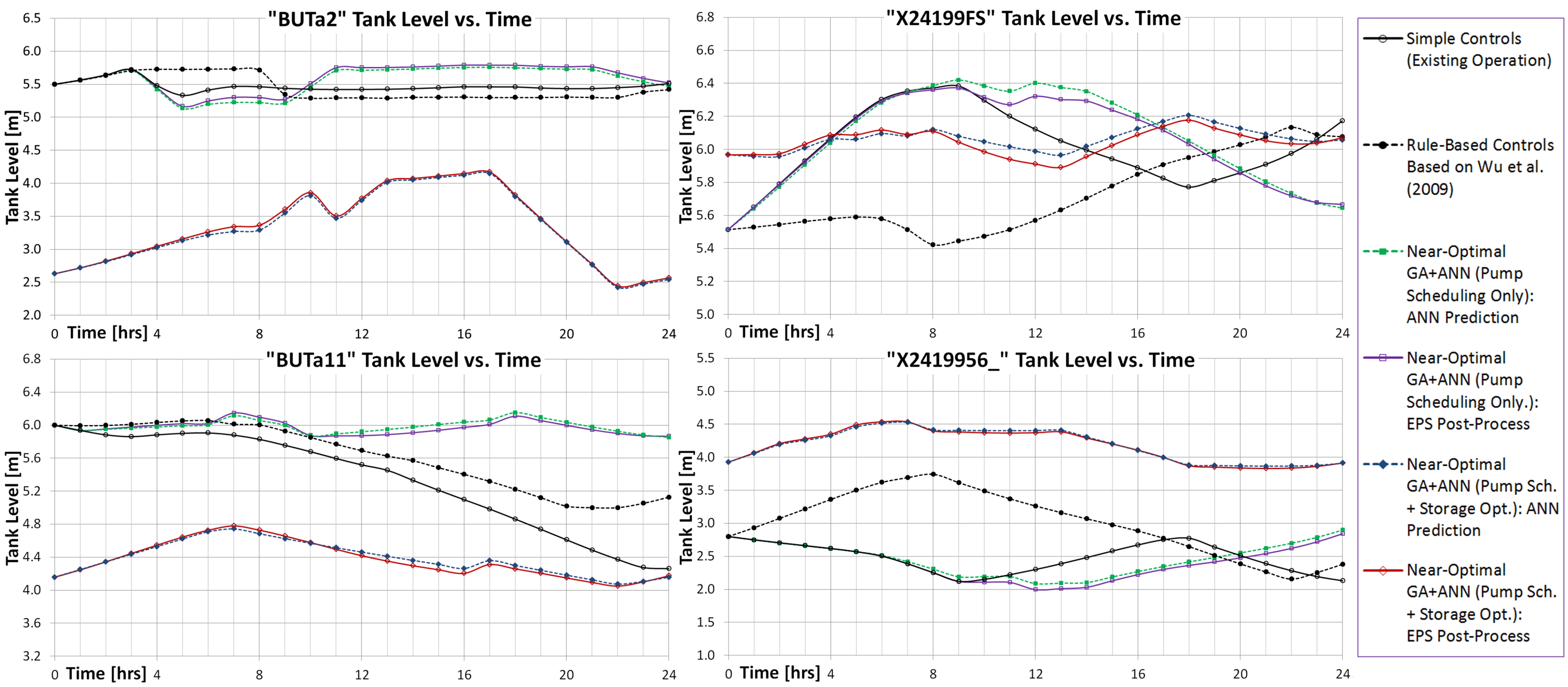}
    \caption{Storage level variations over $24$ hours for $4$ operation scenarios. For GA+ANN results, both meta-model prediction and hydraulic model post-processing results are shown. The post-processing results are generated with EPANET using a simulation time step of $1$ hour.} \label{figure5}
\end{figure}

Fig. \ref{figure4} (a) shows the difference between the initial and the final tank levels for the two GA+ANN scenarios where the shaded area represents the feasible space $|P_{{\rm T},j} (24) - {{\rm T},j} (0)| \leq 10 {\rm cm}$. The energy and cost of both scenarios are plotted in panel (b) and compared with those of the current operation. The solution of pump scheduling and storage optimization is obtained with $\bar{F} = 0.4199$ and $\bar{G} = 0.0000$ (no violation), resulting in saving of around \textlira 130 per day. However, the result of sole pump scheduling is obtained with $\bar{F} = 0.4687$ and $\bar{G} = 0.1966$, showing some violation of the periodicity constraints depicted in Fig. \ref{figure4} (a), and less saving of \textlira 50 per day. Although this result after $5,000$ GA generations does not necessarily mean that no feasible solution exists for the latter case, nevertheless it shows that starting from non-optimal initial conditions of the existing operation introduces difficulties to the GA optimizer to find a feasible solution with significant cost reduction. This demonstrates the importance of a proper choice of the initial combination of tank levels that would be able to recover in $24$ hours, and also explains some of the difficulties faced in the previous pump scheduling studies in \citet{Wu2012,Wu2012a}. Fig. \ref{figure5} compares the tank level variations for the four operation scenarios. For the GA+ANN solutions, both meta-model predictions (used within GA) and hydraulic model post-processing results are illustrated. The meta-model accuracy was validated with the accumulated errors of less than $\pm 5 {\rm cm}$ per $24$ hours, which is a significant improvement over the $20-30 {\rm cm}$ errors of earlier development (\citet{Behandish2012} and \citet{Wu2012,Wu2012a}). It is observed that when the initial tank levels are confined to the existing operation values, the GA+ANN results do not deviate much from the existing operation over the first few hours. The solution with concurrent pump scheduling and storage optimization, on the other hand, utilizes a larger fraction of tank capacities and recovers all of the constrained tank levels to $\pm 10 {\rm cm}$ of initial values. The saving of around $835 {\rm KWH}$ (\textlira 130) per day corresponds to an annual saving of around $300 {\rm MWH}$ (\textlira 480,000), which is significant for a DMZ system of this size.

\section{Conclusions} \label{sec_con}

The study has demonstrated that in addition to the pump scheduling policy, the decisions on tank operation range can play a significant role in water distribution operation cost and storage utilization, and to guarantee the repeatability of the operation policy with predictable behavior over subsequent cycles. The GA+ANN algorithm is generalized to represent a wide range of complex systems and their requirements for pump and valve operation control, carried out concurrently with the optimization of tank operation ranges. Finally, the set of near-optimal tank levels obtained in this offline optimization with a typical demand profile can be useful information for the implementation of the real-time pump operation optimization.

\section*{References}

\bibliography{BENT-TR-14-04}

\newcommand{\noopsort}[1]{} \newcommand{\printfirst}[2]{#1}
  \newcommand{\singleletter}[1]{#1} \newcommand{\switchargs}[2]{#2#1}
\begin{thebibliography}{32}
\providecommand{\natexlab}[1]{#1}
\providecommand{\url}[1]{\texttt{#1}}
\providecommand{\urlprefix}{URL }
\expandafter\ifx\csname urlstyle\endcsname\relax
  \providecommand{\doi}[1]{doi:\discretionary{}{}{}#1}\else
  \providecommand{\doi}[1]{doi:\discretionary{}{}{}\begingroup
  \urlstyle{rm}\url{#1}\endgroup}\fi
\providecommand{\bibinfo}[2]{#2}

\bibitem[{{EPA}(2012)}]{EPA2012}
\bibinfo{author}{{EPA}}, \bibinfo{title}{Energy Efficiency for Water and
  Wastewater Utilities},
  \urlprefix\url{water.epa.gov/infrastructure/sustain/energyefficiency.cfm},
  \bibinfo{year}{2012}.

\bibitem[{Mackle et~al.(1995)Mackle, Savic, and Walters}]{Mackle1995}
\bibinfo{author}{G.~Mackle}, \bibinfo{author}{G.~A. Savic},
  \bibinfo{author}{G.~A. Walters}, \bibinfo{title}{Application of Genetic
  Algorithms to Pump SCheduling for Water Supply}, in: \bibinfo{booktitle}{1st
  International Conference on Genetic Algorithms in Engineering Systems:
  Innovations and Applications ({GALESIA})}, \bibinfo{organization}{IET},
  \bibinfo{pages}{400--405}, \bibinfo{year}{1995}.

\bibitem[{Beckwith and Wong(1995)}]{Beckwith1995}
\bibinfo{author}{S.~P. Beckwith}, \bibinfo{author}{K.~P. Wong},
  \bibinfo{title}{A Genetic Algorithm Approach for Electric Pump Scheduling in
  Water Supply Systems}, in: \bibinfo{booktitle}{{IEEE} International
  Conference on Evolutionary Computation}, vol.~\bibinfo{volume}{1},
  \bibinfo{organization}{IEEE}, \bibinfo{pages}{21}, \bibinfo{year}{1995}.

\bibitem[{Engelbrecht and Haarhoff(1996)}]{Engelbrecht1996}
\bibinfo{author}{R.~J. Engelbrecht}, \bibinfo{author}{J.~Haarhoff},
  \bibinfo{title}{Optimization of Variable-Speed Centrifugal Pump Operation
  with a Genetic Algorithm}, in: \bibinfo{booktitle}{The 3rd International
  Conference on Computer Methods and Water Resources ({CMWR III})},
  \bibinfo{year}{1996}.

\bibitem[{Nitivattananon et~al.(1996)Nitivattananon, Sadowski, and
  Quimpo}]{Nitivattananon1996}
\bibinfo{author}{V.~Nitivattananon}, \bibinfo{author}{E.~Sadowski},
  \bibinfo{author}{R.~Quimpo}, \bibinfo{title}{Optimization of Water Supply
  System Operation}, \bibinfo{journal}{Journal of Water Resources Planning and
  Management} \bibinfo{volume}{122}~(\bibinfo{number}{5})
  (\bibinfo{year}{1996}) \bibinfo{pages}{374--384}.

\bibitem[{De~Schaetzen et~al.(1998)De~Schaetzen, Savic, and
  Walters}]{DeSchaetzen1998}
\bibinfo{author}{W.~B.~F. De~Schaetzen}, \bibinfo{author}{D.~A. Savic},
  \bibinfo{author}{G.~A. Walters}, \bibinfo{title}{A Genetic Algorithm Approach
  to Pump Scheduling in Water-Supply Systems}, in:
  \bibinfo{booktitle}{Hydroinformatics}, vol.~\bibinfo{volume}{98},
  \bibinfo{pages}{897--900}, \bibinfo{year}{1998}.

\bibitem[{Wu et~al.(2001)Wu, Boulos, de~Schaetzen, Orr, Moore, and
  Soft}]{Wu2001}
\bibinfo{author}{Z.~Wu}, \bibinfo{author}{P.~Boulos},
  \bibinfo{author}{W.~de~Schaetzen}, \bibinfo{author}{C.-H. Orr},
  \bibinfo{author}{M.~Moore}, \bibinfo{author}{M.~Soft}, \bibinfo{title}{Using
  Genetic Algorithms for Water Distribution System Optimization}, in:
  \bibinfo{booktitle}{Proceedings of the {ASCE} Environmental and Water
  Resources Institute's ({EWRI}'s) World Water \& Environmental Resource
  Congress}, \bibinfo{pages}{20--24}, \bibinfo{year}{2001}.

\bibitem[{Kelner and L{\'e}onard(2003)}]{Kelner2003}
\bibinfo{author}{V.~Kelner}, \bibinfo{author}{O.~L{\'e}onard},
  \bibinfo{title}{Optimal Pump Scheduling for Water Supply Using Genetic
  Algorithms}, in: \bibinfo{booktitle}{International Congress on Evolutionary
  Methods for Design, Optimization and Control with Applications to Industrial
  Problems ({EUROGEN})}, \bibinfo{year}{2003}.

\bibitem[{Van~Zyl et~al.(2004)Van~Zyl, Savic, and Walters}]{Van2004}
\bibinfo{author}{J.~E. Van~Zyl}, \bibinfo{author}{D.~A. Savic},
  \bibinfo{author}{G.~A. Walters}, \bibinfo{title}{Operational Optimization of
  Water Distribution Systems Using a Hybrid Genetic Algorithm},
  \bibinfo{journal}{Journal of Water Resources Planning and Management}
  \bibinfo{volume}{130}~(\bibinfo{number}{2}) (\bibinfo{year}{2004})
  \bibinfo{pages}{160--170}.

\bibitem[{Savic et~al.(1997)Savic, Walters, and Schwab}]{Savic1997}
\bibinfo{author}{D.~A. Savic}, \bibinfo{author}{G.~A. Walters},
  \bibinfo{author}{M.~Schwab}, \bibinfo{title}{Multiobjective Genetic
  Algorithms for Pump Scheduling in Water Supply}, in:
  \bibinfo{booktitle}{Evolutionary Computing Lecture Notes in Computer
  Science}, vol. \bibinfo{volume}{1305}, \bibinfo{publisher}{Springer},
  \bibinfo{pages}{227--235}, \bibinfo{year}{1997}.

\bibitem[{Sotelo et~al.(2002)Sotelo, von L{\"u}cken, and
  Bar{\'a}n}]{Sotelo2002}
\bibinfo{author}{A.~Sotelo}, \bibinfo{author}{C.~von L{\"u}cken},
  \bibinfo{author}{B.~Bar{\'a}n}, \bibinfo{title}{Multiobjective Evolutionary
  Algorithms in Pump Scheduling Optimisation}, in:
  \bibinfo{booktitle}{Proceedings of the 3rd International Conference on
  Engineering Computational Technology}, \bibinfo{organization}{Civil-Comp
  press}, \bibinfo{pages}{175--176}, \bibinfo{year}{2002}.

\bibitem[{Bar{\'a}n et~al.(2005)Bar{\'a}n, von L{\"u}cken, and
  Sotelo}]{Baran2005}
\bibinfo{author}{B.~Bar{\'a}n}, \bibinfo{author}{C.~von L{\"u}cken},
  \bibinfo{author}{A.~Sotelo}, \bibinfo{title}{Multi-Objective Pump Scheduling
  Optimisation Using Evolutionary Strategies}, \bibinfo{journal}{Advances in
  Engineering Software} \bibinfo{volume}{36}~(\bibinfo{number}{1})
  (\bibinfo{year}{2005}) \bibinfo{pages}{39--47}.

\bibitem[{L{\'o}pez-Ib{\'a}{\~n}ez et~al.(2005)L{\'o}pez-Ib{\'a}{\~n}ez,
  Devi~Prasad, and Paechter}]{Lopez2005}
\bibinfo{author}{M.~L{\'o}pez-Ib{\'a}{\~n}ez},
  \bibinfo{author}{T.~Devi~Prasad}, \bibinfo{author}{B.~Paechter},
  \bibinfo{title}{Multi-Objective Optimisation of the Pump Scheduling Problem
  Using {SPEA}2}, in: \bibinfo{booktitle}{{IEEE} Congress on Evolutionary
  Computation}, vol.~\bibinfo{volume}{1}, \bibinfo{organization}{IEEE},
  \bibinfo{pages}{435--442}, \bibinfo{year}{2005}.

\bibitem[{Wang et~al.(2009)Wang, Chang, and Chen}]{Wang2009}
\bibinfo{author}{J.-Y. Wang}, \bibinfo{author}{T.-P. Chang},
  \bibinfo{author}{J.-S. Chen}, \bibinfo{title}{An Enhanced Genetic Algorithm
  for Bi-Objective Pump Scheduling in Water Supply}, \bibinfo{journal}{Expert
  Systems with Applications} \bibinfo{volume}{36}~(\bibinfo{number}{7})
  (\bibinfo{year}{2009}) \bibinfo{pages}{10249--10258}.

\bibitem[{von L{\"u}cken et~al.(2004)von L{\"u}cken, Bar{\'a}n, and
  Sotelo}]{Von2004}
\bibinfo{author}{C.~von L{\"u}cken}, \bibinfo{author}{B.~Bar{\'a}n},
  \bibinfo{author}{A.~Sotelo}, \bibinfo{title}{Pump Scheduling Optimization
  Using Asynchronous Parallel Evolutionary Algorithms},
  \bibinfo{journal}{{CLEI} Electronic Journal}
  \bibinfo{volume}{7}~(\bibinfo{number}{2}).

\bibitem[{Wu and Zhu(2009)}]{Wu2009a}
\bibinfo{author}{Z.~Y. Wu}, \bibinfo{author}{Q.~Zhu}, \bibinfo{title}{Scalable
  Parallel Computing Framework for Pump Scheduling Optimization}, in:
  \bibinfo{booktitle}{World Water and Environmental Resource Congress},
  \bibinfo{year}{2009}.

\bibitem[{Wu and Behandish(2012{\natexlab{a}})}]{Wu2012}
\bibinfo{author}{Z.~Y. Wu}, \bibinfo{author}{M.~Behandish},
  \bibinfo{title}{Comparing Methods of Parallel Genetic Optimization for Pump
  Scheduling Using Hydraulic Model and {GPU}-Based {ANN} Meta-Model}, in:
  \bibinfo{booktitle}{14th Water Distribution Systems Analysis Conference
  ({WDSA})}, \bibinfo{organization}{Engineers Australia},
  \bibinfo{pages}{1088}, \bibinfo{year}{2012}{\natexlab{a}}.

\bibitem[{Wu and Behandish(2012{\natexlab{b}})}]{Wu2012a}
\bibinfo{author}{Z.~Y. Wu}, \bibinfo{author}{M.~Behandish},
  \bibinfo{title}{Real-Time Pump Scheduling Using Genetic Algorithm and
  Artificial Neural Network Based on Graphics Processing Unit}, in:
  \bibinfo{booktitle}{14th Water Distribution Systems Analysis Conference
  ({WDSA})}, \bibinfo{organization}{Engineers Australia},
  \bibinfo{pages}{1088}, \bibinfo{year}{2012}{\natexlab{b}}.

\bibitem[{Rao and Alvarruiz(2007)}]{Rao2007}
\bibinfo{author}{Z.~Rao}, \bibinfo{author}{F.~Alvarruiz}, \bibinfo{title}{Use
  of an Artificial Neural Network to Capture the Domain Knowledge of a
  Conventional Hydraulic Simulation Model}, \bibinfo{journal}{Journal of
  Hydroinformatics} \bibinfo{volume}{9}~(\bibinfo{number}{1})
  (\bibinfo{year}{2007}) \bibinfo{pages}{15--24}.

\bibitem[{Broad et~al.(2005)Broad, Dandy, and Maier}]{Broad2005}
\bibinfo{author}{D.~R. Broad}, \bibinfo{author}{G.~C. Dandy},
  \bibinfo{author}{H.~R. Maier}, \bibinfo{title}{Water Distribution System
  Optimization Using Metamodels}, \bibinfo{journal}{Journal of Water Resources
  Planning and Management} \bibinfo{volume}{131}~(\bibinfo{number}{3})
  (\bibinfo{year}{2005}) \bibinfo{pages}{172--180}.

\bibitem[{Tabach et~al.(2007)Tabach, Lancelot, Shahrour, and
  Najjar}]{Tabach2007}
\bibinfo{author}{E.~E. Tabach}, \bibinfo{author}{L.~Lancelot},
  \bibinfo{author}{I.~Shahrour}, \bibinfo{author}{Y.~Najjar},
  \bibinfo{title}{Use of Artificial Neural Network Simulation Metamodelling to
  Assess Groundwater Contamination in a Road Project},
  \bibinfo{journal}{Mathematical and Computer Modelling}
  \bibinfo{volume}{45}~(\bibinfo{number}{7}) (\bibinfo{year}{2007})
  \bibinfo{pages}{766--776}.

\bibitem[{Jamieson et~al.(2007)Jamieson, Shamir, Martinez, and
  Franchini}]{Jamieson2007}
\bibinfo{author}{D.~Jamieson}, \bibinfo{author}{U.~Shamir},
  \bibinfo{author}{F.~Martinez}, \bibinfo{author}{M.~Franchini},
  \bibinfo{title}{Conceptual design of a Generic, Real-time, Near-Optimal
  Control System for Water-Distribution Networks}, \bibinfo{journal}{Journal of
  Hydroinformatics} \bibinfo{volume}{9}~(\bibinfo{number}{1})
  (\bibinfo{year}{2007}) \bibinfo{pages}{3--14}.

\bibitem[{Rao and Salomons(2007)}]{Rao2007a}
\bibinfo{author}{Z.~Rao}, \bibinfo{author}{E.~Salomons},
  \bibinfo{title}{Development of a Real-Time, Near-Optimal Control Process for
  Water-Distribution Networks}, \bibinfo{journal}{Journal of Hydroinformatics}
  \bibinfo{volume}{9}~(\bibinfo{number}{1}) (\bibinfo{year}{2007})
  \bibinfo{pages}{25--37}.

\bibitem[{Salomons et~al.(2007)Salomons, Goryashko, Shamir, Rao, and
  Alvisi}]{Salomons2007}
\bibinfo{author}{E.~Salomons}, \bibinfo{author}{A.~Goryashko},
  \bibinfo{author}{U.~Shamir}, \bibinfo{author}{Z.~Rao},
  \bibinfo{author}{S.~Alvisi}, \bibinfo{title}{Optimizing the Operation of the
  Haifa-A Water-DistributionNetwork}, \bibinfo{journal}{Journal of
  Hydroinformatics} \bibinfo{volume}{9}~(\bibinfo{number}{1})
  (\bibinfo{year}{2007}) \bibinfo{pages}{51--64}.

\bibitem[{Martinez et~al.(2007)Martinez, Hernandez, Alonso, Rao, and
  Alvisi}]{Martinez2007}
\bibinfo{author}{F.~Martinez}, \bibinfo{author}{V.~Hernandez},
  \bibinfo{author}{J.~Alonso}, \bibinfo{author}{Z.~Rao},
  \bibinfo{author}{S.~Alvisi}, \bibinfo{title}{Optimizing the Operation of the
  Valencia Water-Distribution Network}, \bibinfo{journal}{Journal of
  Hydroinformatics} \bibinfo{volume}{9}~(\bibinfo{number}{1})
  (\bibinfo{year}{2007}) \bibinfo{pages}{65--78}.

\bibitem[{Shamir and Salomons(2008)}]{Shamir2008}
\bibinfo{author}{U.~Shamir}, \bibinfo{author}{E.~Salomons},
  \bibinfo{title}{Optimal Real-Time Operation of Urban Water Distribution
  Systems Using Reduced Models}, \bibinfo{journal}{Journal of Water Resources
  Planning and Management} \bibinfo{volume}{134}~(\bibinfo{number}{2})
  (\bibinfo{year}{2008}) \bibinfo{pages}{181--185}.

\bibitem[{Behandish and Wu(2012)}]{Behandish2012}
\bibinfo{author}{M.~Behandish}, \bibinfo{author}{Z.~Y. Wu},
  \bibinfo{title}{{GPU}-based Artificial Neural Network Configuration and
  Training for Water Distribution System Analysis}, in:
  \bibinfo{booktitle}{ASCE Annual World Environmental and Water Resources
  Congress}, \bibinfo{address}{Albuquerque, NM, U.S.A.}, \bibinfo{year}{2012}.

\bibitem[{Wu and Eftekharian(2011)}]{Wu2011a}
\bibinfo{author}{Z.~Y. Wu}, \bibinfo{author}{A.~A. Eftekharian},
  \bibinfo{title}{Parallel Artificial Neural Network Using {CUDA}-Enabled {GPU}
  for Extracting Hydraulic Domain Knowledge of Large Water Distribution
  Systems}, in: \bibinfo{booktitle}{ASCE Annual World Environmental and Water
  Resources Congress}, \bibinfo{year}{2011}.

\bibitem[{Behandish and Wu(2013)}]{Behandish2013}
\bibinfo{author}{M.~Behandish}, \bibinfo{author}{Z.~Y. Wu},
  \bibinfo{title}{Generalized {GPU}-based Artificial Neural Network Surrogate
  Model for Extended Period Hydraulic Simulation}, in: \bibinfo{booktitle}{ASCE
  Annual World Environmental and Water Resources Congress},
  \bibinfo{year}{2013}.

\bibitem[{NVIDIA(2012)}]{CUDA2012}
\bibinfo{author}{NVIDIA}, \bibinfo{title}{CUDA C Programming Guide},
  \urlprefix\url{docs.nvidia.com/cuda/pdf/CUDA\_C\_Programming\_Guide.pdf},
  \bibinfo{year}{2012}.

\bibitem[{Mundo and Yan(2007)}]{Mundo2007}
\bibinfo{author}{D.~Mundo}, \bibinfo{author}{H.-S. Yan},
  \bibinfo{title}{Kinematic Optimization of Ball-Screw Transmission
  Mechanisms}, \bibinfo{journal}{Mechanism and Machine Theory}
  \bibinfo{volume}{42}~(\bibinfo{number}{1}) (\bibinfo{year}{2007})
  \bibinfo{pages}{34--47}.

\bibitem[{Wu et~al.(2009)Wu, Woodward, and Allen}]{Wu2009}
\bibinfo{author}{Z.~Y. Wu}, \bibinfo{author}{K.~Woodward},
  \bibinfo{author}{T.~Allen}, \bibinfo{title}{Pump scheduling study on a demand
  monitoring zone}, in: \bibinfo{booktitle}{International Conference on
  Computing and Control for Water Industry (CCWI)}, \bibinfo{year}{2009}.

\end{thebibliography}
\bibliographystyle{elsarticle-num-names}

\end{document}